\documentclass[runningheads]{llncs}
\usepackage{graphicx}
\usepackage{amssymb}
\usepackage{algorithm}
\usepackage{algpseudocode}
\usepackage{caption}
\usepackage{subcaption}
\usepackage{multirow}
\usepackage{amsmath}
\usepackage{hyperref}
\usepackage{comment}
\begin{document}
 \title{A Deep Reinforcement Learning Approach to Battery Management in Dairy Farming via Proximal Policy Optimization}
%
%
\author{Nawazish Ali\inst{1} \and
Rachael Shaw\inst{2} \and
Karl Mason\inst{1}}
\authorrunning{N. Ali et al.}
%
\institute{School of Computer Science, University of Galway, Galway, H91 TK33, Ireland  \and
Atlantic Technological University, Galway, H91 T8NW, Ireland
}

%
\maketitle              
\begin{abstract}
Dairy farms consume a significant amount of electricity for their operations, and this research focuses on enhancing energy efficiency and minimizing the impact on the environment in the sector by maximizing the utilization of renewable energy sources. This research investigates the application of Proximal Policy Optimization (PPO), a deep reinforcement learning algorithm (DRL), to enhance dairy farming battery management. We evaluate the algorithm's effectiveness based on its ability to reduce reliance on the electricity grid, highlighting the potential of DRL to enhance energy management in dairy farming. Using real-world data our results demonstrate how the PPO approach outperforms Q-learning by 1.62\% for reducing electricity import from the grid.
This significant improvement highlights the potential of the Deep Reinforcement Learning algorithm for improving energy efficiency and sustainability in dairy farms.

\end{abstract}
\section{Introduction}
The continuous growth of the global population has escalated the demand for dairy products, positioning dairy farming as an important sector of agriculture\cite{dairy_growth}. The OECD-FAO Agricultural Outlook 2020–2029 predicts a 1.6\% annual increase in milk production to 997 metric tons by 2029. This increased demand has increased milk production and expanded the worldwide export of dairy products\cite{dairy_export}. Dairy farms consume a significant amount of electricity for different operations, from milking to cooling and storage \cite{dairy_operations}. The increase in milk production also increases the farm's electricity demand. Due to the growing electricity demand, the dairy farm industry needs to focus more on enhancing efficiency and sustainability in their operations. This necessitates innovative approaches to manage the energy-intensive processes involved in dairy farming to ensure sustainability.\\

With increasing demand for electricity, in recent years there has been a significant increase in the integration of renewable energy sources for sustainability in dairy farming\cite{ahdb_renewable_energy}. The adoption of renewable energy shows the industry's focus on reducing carbon footprints and adopting sustainable energy sources. However, the intermittent nature of renewable energy generation poses a significant challenge. This variability emphasizes the need for efficient energy management solutions to mitigate the variations between energy generation and consumption.\\
Recent advances in Artificial Intelligence (AI) and, specifically, DRL \cite{ai_advancement}, offer a promising path to address the challenges mentioned above, by integrating renewable energy and managing batteries in dairy farming. DRL, a subset of AI, excels in making decisions in complex, uncertain environments by learning optimal actions through trial and error. This capability makes DRL ideal for optimizing energy usage and storage in fluctuating renewable energy supplies. By utilizing DRL algorithms, dairy farms can dynamically control their energy consumption and storage based on real-time data, optimizing renewable energy use and enhancing overall operational efficiency.\\
The main objective of this paper is to explore the application of PPO \cite{ppo}, a state-of-the-art DRL algorithm, in optimizing battery management for dairy farming operations. 
This paper highlights the potential of PPO to transform energy management systems in dairy farming, contributing to the sector's long-term sustainability and resilience by enhancing global energy transitions. \\

The main contributions of this research are highlighted below: 
\begin{itemize}
\item In contrast to existing approaches this research is the first to apply Deep Reinforcement Learning for battery management in the dairy farming sector.
  \item To compare the performance of the DRL algorithm with traditional rule-based and Q-learning methods. 
  \item Analyze the policy learned by the DRL algorithm for controlling the battery. 
\end{itemize}

\section{Background and Related Research}

\subsection{Reinforcement Learning}
Reinforcement Learning(RL) is an important component of Artificial Intelligence in which an agent learns to make decisions by interacting with an environment. The RL agent learns a policy $\pi$ which it believes will maximize the accumulated reward. This is achieved through an iterative process of exploration, where the agent observes the possible states from the environment \textbf{\textit{S}}, performs an action \textbf{\textit{A}}, and gets a reward \textbf{\textit{R}} from the environment, that leads to a transition to a new state \textbf{\textit{S'}}. This interaction is commonly represented as a Markov Decision Process (MDP), characterized by the tuple (\textbf{\textit{S, A, P, R, $\gamma$}}), where \textbf{\textit{P}} refers to the probability of transition of the state and \textbf{\textit{$\gamma$}} represents the discount factor that balances immediate and future reward.
RL optimizes the policy based on the state and action value function denoted as  \textbf{\textit{Q(s, a)}}, which represents the expected reward by exploring the state and taking action by following the policy $\pi$. The state and action-value function is presented in Equation \ref{bellman}

\begin{equation}
Q^\pi(s,a) = \mathbb{E} \left[ R_{t+1} + \gamma Q^\pi(S_{t+1}, A_{t+1}) | S_t = s, A_t = a \right]
\label{bellman}
\end{equation}

Equation \ref{bellman} shows the expected future reward $R_{t+1}$ after taking action $a$ in state $s$, the discounted future reward as represented by $\gamma Q^\pi(S_{t+1}, A_{t+1})$ given the current state-action pair.

\subsection{Proximal Policy Optimization (PPO)}

PPO is an advanced RL algorithm that is developed to enhance the stability and efficiency of policy gradient methods in reinforcement learning. It addresses the issue of large policy updates, which can lead to reduced performance, by introducing a mechanism to limit policy updates to ensure a more stable learning process. PPO utilizes a clipping objective function to limit the policy updates to a limited range, ensuring that the policy does not deviate too far from the previous policy. The objective function of the PPO is expressed in Equation \ref{objective_function}

\begin{equation}
L^{CLIP}(\theta) = \hat{\mathbb{E}}_t \left[ \min\left(r_t(\theta) \hat{A}_t, \text{clip}\left(r_t(\theta), 1-\delta, 1+\delta\right) \hat{A}_t\right) \right]
\label{objective_function}
\end{equation}

In Equation \ref{objective_function} $r_t(\theta)$ represents the ratio of the probability of the new policy over the old policy, $\hat{A}_t$ is an estimator of the advantage at time t, and ($\delta$) is a hyperparameter that defines the limits for clipping, thus limiting the range of policy updates. PPO is an advanced approach for solving the policy optimization problem and managing complex environments, along with its outstanding performance across different applications.

\subsection{Related Work}
Researchers are investigating a wide range of approaches to improve energy utilization in the context of battery management, which has gained significant attention in various applications. Various rule-based battery management techniques, such as Maximizing Self-Consumption (MSC) and Time of Use (TOU), as well as optimization methods, have been widely used in different settings \cite{braun,luthander,gitizadeh,hassan}. These methods optimize the utilization of locally produced solar power and take advantage of off-peak electricity prices to efficiently use batteries. 

However, the emergence of AI and RL has encouraged the way for more sophisticated approaches to battery management. RL, in particular, is well-suited for developing optimal solutions that involve engaging with and learning from the environment. This method is considered a potential strategy for enhancing energy management by leveraging the capability to gain information through interaction with the environment. Various RL techniques have been applied to optimize battery management across different application scenarios.

For example, Foruzan et al. introduced RL for the management of energy in microgrids, demonstrating its adaptability to changing energy needs and improving energy efficiency \cite{foruzan}. Guan et al. developed an RL-based solution for domestic energy storage control, effectively reducing electricity costs by optimizing charging and discharge strategies \cite{guan}. Cao et al. proposed a DRL method for battery charging and discharging, effectively handling the uncertainty of the power price and improving the accuracy of the degradation model \cite{cao}.

Yu et al. used the deep-deterministic policy gradient (DDPG) to minimize electricity costs, achieving significant energy savings \cite{yu2019deep}. Wei et al. proposed DDPG for fast charging of lithium-ion batteries, considering various constraints such as battery temperature and charging speed \cite{wei}. Liu et al. explored DRL for optimizing energy management in the home, demonstrating its performance over traditional methods to improve energy efficiency \cite{liu}. Additionally, Cheng et al. introduced a periodic deterministic policy gradient algorithm (PDPG) to schedule multibattery energy storage systems, achieving significant power cost reductions \cite{cheng}. Huang et al. introduced PPO for optimizing the capacity scheduling of solar battery systems, enhancing battery safety \cite{huang}. Paudel et al. used the Markov Decision Process (MDP) framework to efficiently manage battery storage systems, considering fluctuations in electricity prices \cite{paudel}. Ali et al. explored battery management strategies for dairy farms, employing both rule-based and Q-learning approaches \cite{ali2024reinforcement}. Their research demonstrated a notable reduction in grid electricity consumption by up to 10.64\% through the application of Q-learning. Although its investigation did not extend to deep reinforcement learning (DRL) for dairy farm settings, this study is an extension of their work by integrating DRL techniques for enhanced battery management within dairy farm operations.

Despite these advancements in RL techniques for battery management, there is a notable gap in the literature as DRL has not yet been applied to battery management in the context of dairy farming. This research aims to address this gap by utilizing the DRL methods to optimize battery management in the dairy farming industry.

\section{Methodology}
\subsection{Environment Design}

\begin{figure}[h]
\centering
\includegraphics[width=0.95\textwidth]{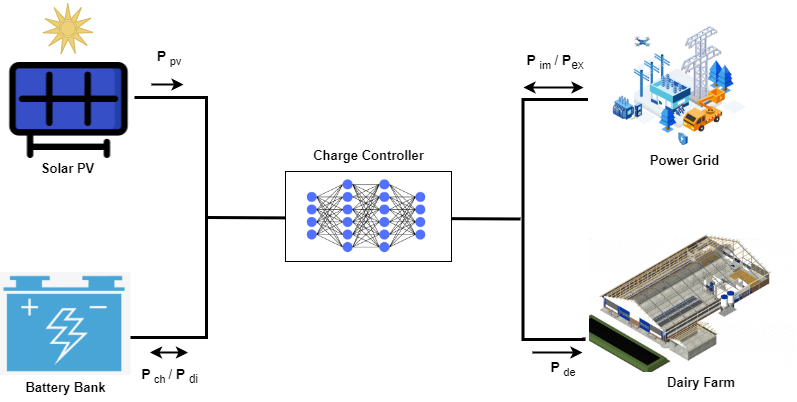}
\caption{Overview of the system environment.} \label{grid}
\end{figure}

The study's environment, shown in Figure \ref{grid}, includes solar PV, a Tesla Powerwall 2.0 (13.5 kWh capacity, 5 kW charge/discharge rate)\cite{teslapowerwall}, a power grid, and a dairy farm. PV electricity can either meet the farm's needs or charge the battery. A controller optimizes battery management based on energy generation, demand, and pricing. The power grid supplies electricity during high demand and low renewable generation periods. The battery system performs peak shaving to meet peak electricity demand.


\subsection{Data Description}
In this study, the dataset used was collected from Finland. The data have information on the farm's electricity consumption, PV generated in the farm, and the price of electricity. The electricity consumption data is collected \cite{fin_data}, which contains information on hourly electricity consumption over one year, with a consumption of 261 MW per year. The PV electricity generation data is collected from the NREL Advisor Model \cite{fin_pv}, which has information on hourly generation over one year with a capacity of 20 KW. Electric price data is collected from the Helsinki electricity supply company website \cite{e_price}, which has dynamic prices that include three different price levels \cite{b_tou}.

\subsection{Deep Reinforcement Learning for Battery Management}
This study applies the PPO algorithm to manage battery storage in a dairy farm environment, focusing on optimizing the use of renewable energy sources. This approach involves an exploration of the defined state and action spaces,  and calculating rewards based on renewable energy availability and electricity pricing. The components of the state space, action space, and reward function are explained below.

The training parameters for the PPO algorithm are, the learning rate was set to 0.003, the exploration rate ($\epsilon$) starting at 1.0, with a decay rate of 0.0001 to mitigate overfitting; and the discount factor ($\gamma$) is established at 0.89, balancing immediate and future rewards. The PPO algorithm uses the clipping parameter ($\delta$), set to 0.2, to moderate the policy update step, ensuring that updates remain within a reasonable range for stable learning. 
 
The optimization process iterates over multiple epochs with a minibatch size of 64, facilitating efficient learning by interacting with the dairy farm environment. The grid search algorithm is used in this work to optimize the best hyperparameters. 

\subsubsection{State Space}
The state space of the dairy farm environment is represented as $S$, which includes all the essential information about the environment of battery management for decision-making. The state space of the environment is represented in Equation \ref{state_space} 

\begin{equation}
S = (hour, SOC, load, PV)
\label{state_space}
\end{equation}

The $(hour)$ represents the time of day, which is important for decision-making related to battery management. By including the time in the state space, the algorithm can learn and apply different strategies depending on the time of day, optimizing energy usage and storage throughout the 24-hour cycle. $(SOC)$ represents the current charge level of the battery system, discretized between 0 and 10 for effective learning. If the SOC is higher, it can be used to meet the farm's energy demand without importing electricity from the grid. The $(load)$ variable represents the current energy demand of the dairy farm. The $(PV)$ indicates the current availability of solar energy, and the availability of PV influences decisions on when to store energy and when to use it directly, for managing the SOC optimally.
\subsubsection{Action Space}
The action space, denoted as $(A)$, comprises discrete actions that the algorithm can take at any given timestep to manage the battery storage. The action space of the algorithm is represented in Equation \ref{action_space}

\begin{equation}
A =  (Charge, Discharge, Idle)
\label{action_space}
\end{equation}

The agent determines the action $(Charge)$ to charge the battery at battery charge rate $(\gamma)$ at a specific time by analyzing the dairy farm's energy demand, PV generation, and electricity prices. The action $(Discharge)$ is chosen to discharge the battery when electricity prices are elevated or when it is necessary to meet the farm's energy demand. The action $(Idle)$ is selected when neither charging nor discharging the battery is deemed optimal.

\subsubsection{Reward}
The reward function, denoted by $(R)$, is determined by calculating the amount of electricity imported from the grid, factoring the electricity price. Equation \ref{reward} provides a mathematical expression for calculating the reward within the battery management environment.

\begin{equation} \label{reward} 
R = \begin{cases}
    -((P_{load} + (\gamma - P_{pv})) \times E_{price}) - Penalty   \text{ if } A = Charge \\
    -((P_{load} - P_{pv}) - \gamma)) \times E_{price}) - Penalty   \text{ if } A = Discharge \\
    -(P_{load} - P_{pv}) \times E_{price}    \text{ if } A = Idle \\
\end{cases}
\end{equation}

$(P_{pv})$ denotes the aggregate power output from the solar panels at a given time instance $(t)$, while $(P_{load})$ determines the electricity demand by the dairy farm. The parameter $(\gamma)$ is defined as the rate at which the battery is charged and discharged measured in kilowatts (kW). $(A)$ denotes the action taken at $(hour)$ and $(E_{price})$ represents the price of electricity at the current timestep. The $(Penalty)$ is the value by which the agent is penalized based on action taken in certain conditions. The detailed equation for determining the penalty is presented in Equation \ref{penalty}

\begin{equation} \label{penalty} 
Penalty = \begin{cases}
    -15 \text{ if } SOC \geq SOC_{\text{max}} \text{ and } A = Charge \\
    -15 \text{ if } SOC \leq SOC_{\text{min}} \text{ and } A = Discharge \\
\end{cases}
\end{equation}

In Equation \ref{penalty} $(SOC)$ is represented as the battery's current state of charge, and the $(SOC_{max})$ represents the battery's maximum charge level. The SOC threshold is set between 15 to 85 percent by setting $(SOC_{min})$ 15\% and $(SOC_{max})$ to 85\%, to enhance both the efficiency and the lifespan of the battery system\cite{battery_university_808}.  The agent is penalized when the battery is fully charged but the agent still tries to Charge it or if the battery is at a minimum level and the agent tries to discharge the battery.\\

The PPO algorithm, outlined in Algorithm \ref{ppo_algorithm}, initializes policy and critic networks and sets hyperparameters. The policy network determines the agent's actions, while the critic network estimates future rewards. The algorithm collects trajectories, calculates an advantage function, and interacts with the environment to obtain rewards. The advantage function assesses each action's benefit. The algorithm then optimizes the policy via gradient ascent and updates the critic network to minimize loss.


\begin{algorithm}[h]
\caption{Proximal Policy Optimization (PPO) Algorithm}
\label{ppo_algorithm}
\begin{algorithmic}[1]

\State Initialize policy parameters $\theta$ with random values
\State Initialize the critic network parameters $\phi$ with random values
\State Set the learning rate $\alpha$, discount factor $\gamma$, and clipping parameter $\epsilon$

\For{iteration $= 1,2,\dots,N$}
    \State Collect set of trajectories by executing the current policy $\pi_{\theta}$ in the environment
    \State Compute rewards and advantage function $\hat{A}_t$ for each time step
    \State Optimize the objective function for a fixed number of epochs:

        \For{each epoch}
            \State Update $\theta$ using gradient ascent to maximize reward by optimizing policy.
        \EndFor

    \State Update the critic network by minimizing the squared loss between predicted and actual returns
\EndFor

\end{algorithmic}
\end{algorithm}

\section{Results and Discussion}
\begin{figure}[H]
\centering
\includegraphics[width=0.95\textwidth]{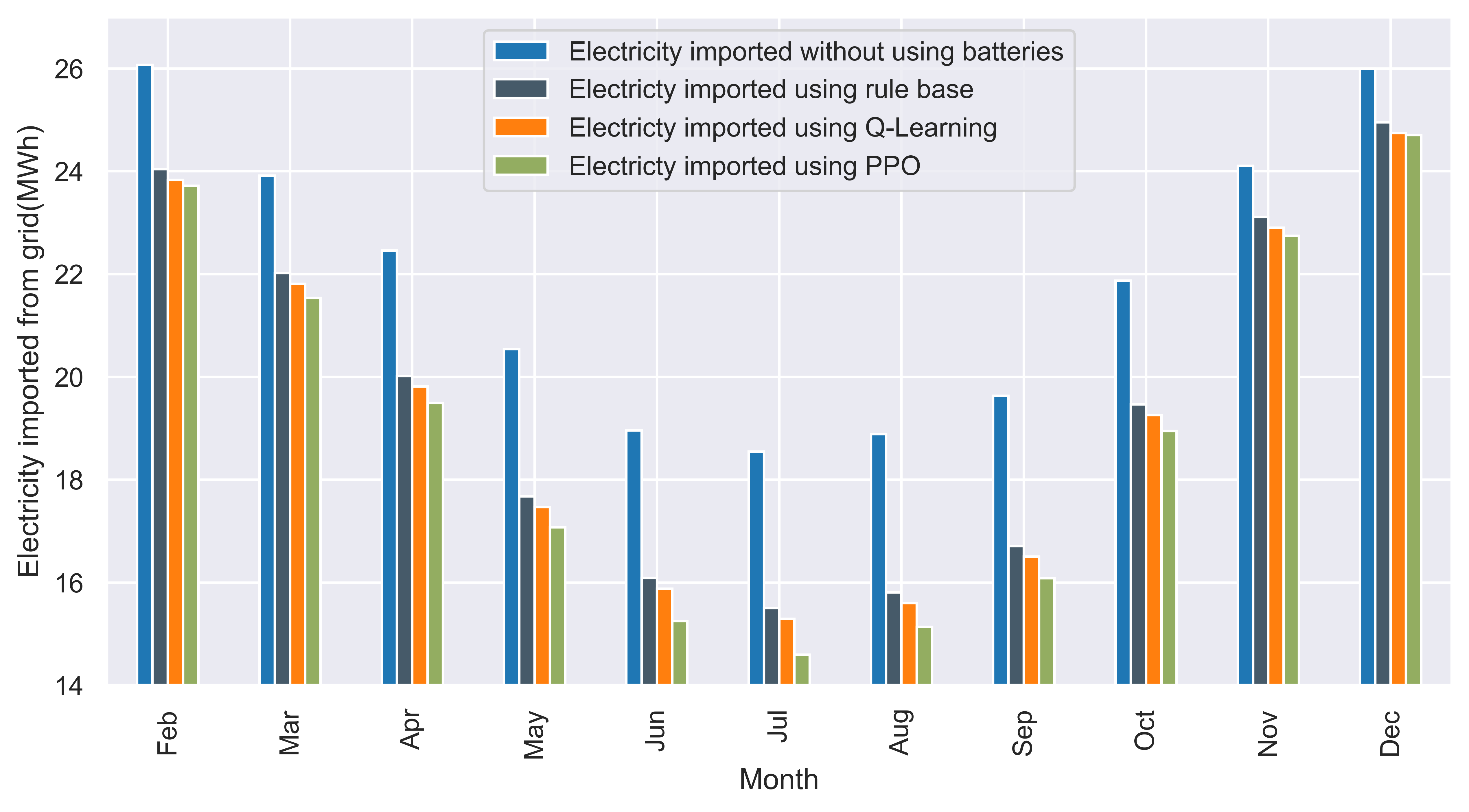}
\caption{Comparison of load imported from the grid by different algorithms.} \label{load_reduction}
\end{figure}

This research utilizes the PPO algorithm to improve battery management in the dairy farming industry, focusing on reducing the amount of electricity imported from the grid. The utilization of PPO exhibited a notable enhancement in efficiently handling energy requirements. Utilizing the PPO algorithm, there is a notable decrease in electricity consumption from the grid by 13.11\% compared to when there is no battery. We compare our algorithm to Q-learning\cite{ali2024reinforcement} and a rule-based\cite{ali2024reinforcement} approach. Our results show an improvement of 1.62\% and 2.56\% respectively . Figure \ref{load_reduction} presents the algorithm results from evaluation results from evaluations over February to December, each month featuring four bars each corresponding to the different methodologies being compared. The findings show that, when compared with Q-learning, our approach significantly enhances electricity savings during the summer months by utilizing high solar energy. However, in winter, as solar generation decreases, the importation of electricity from the grid increases. This trend highlights our algorithm's ability to maximize renewable energy utilization.\\

The algorithm is trained over a month, using data from the 1st to the 30th of January, and tested on the data from Feb to December. The reward function design in this experiment does not constrain the algorithm to select predefined optimal actions. Instead, it strategically penalizes actions that could harm the battery's efficiency, such as charging it when full or discharging it when empty. The agent is not forced to learn favorable actions, instead, it freely explores its environment and determines its policy for maximizing the reward function.
It allows the algorithm to make a more robust and adaptable decision-making policy.

\begin{figure}[H]
\centering
\includegraphics[width=0.95\textwidth]{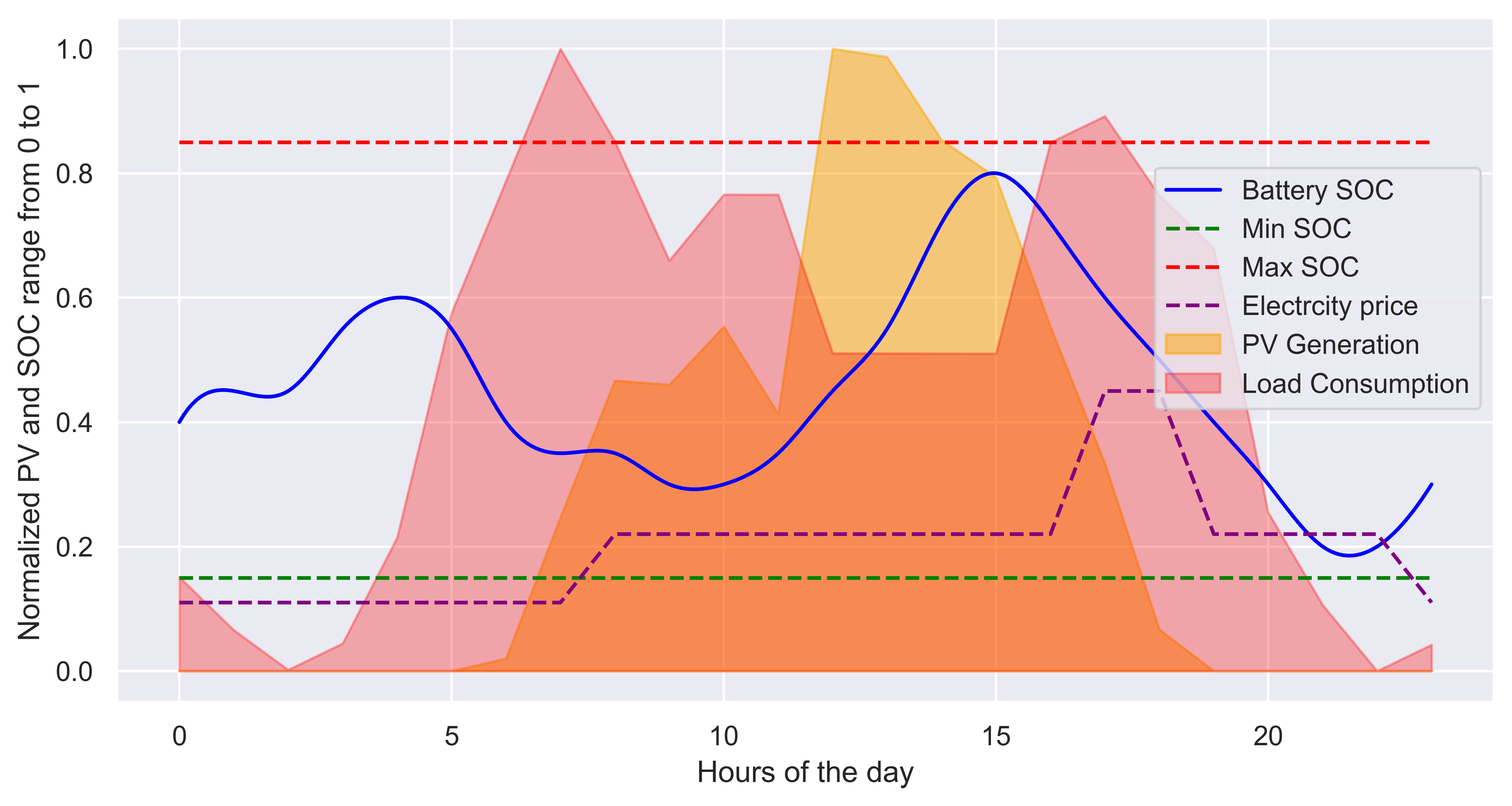}
\caption{Agent behavior for battery charging and discharging during the day.} \label{battery_control}
\end{figure}

\begin{figure}[H]
\centering
\includegraphics[width=0.95\textwidth]{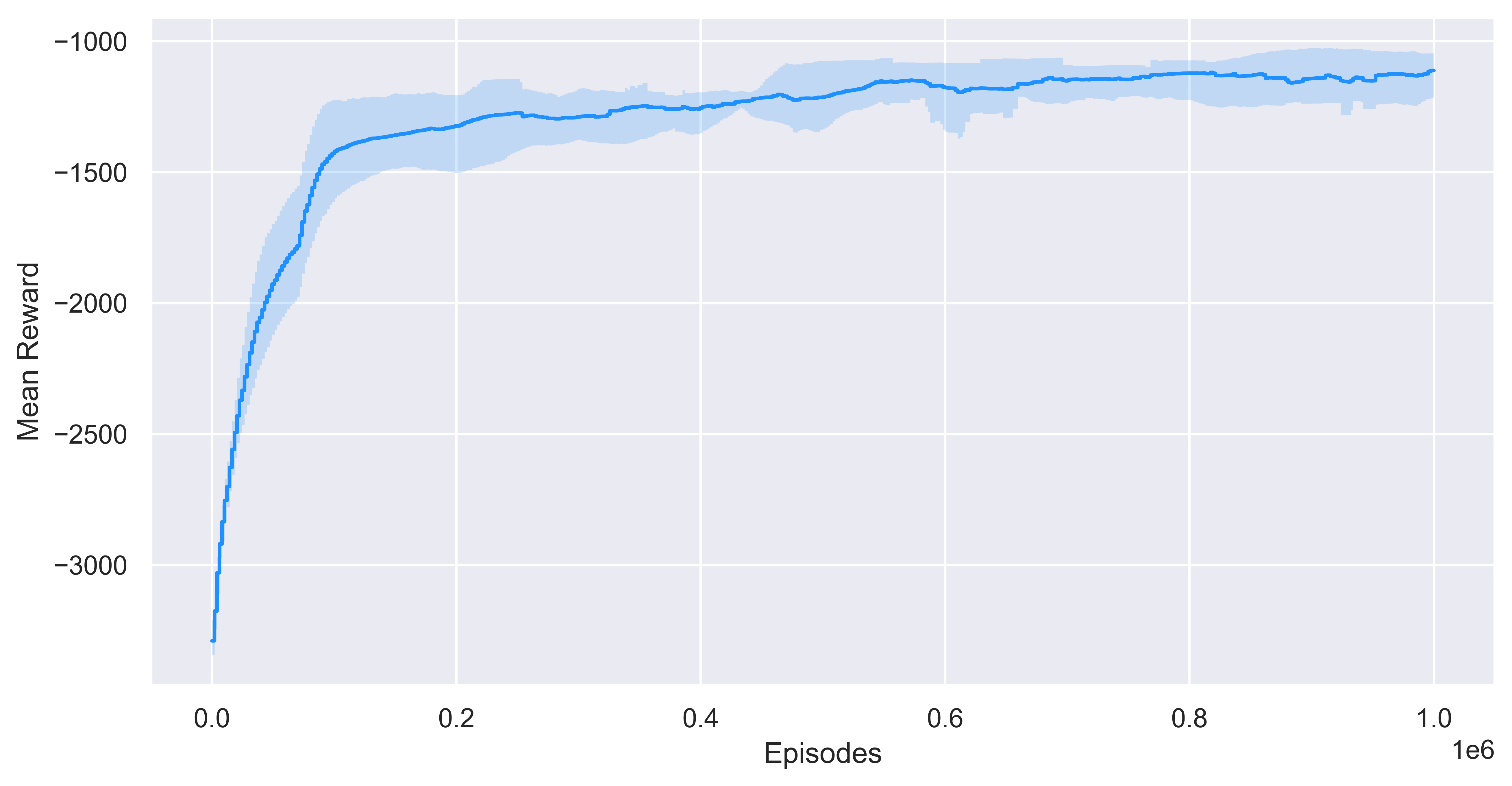}
\caption{Training reward of implemented PPO algorithm} \label{reward}
\end{figure}

Figure \ref{battery_control} shows the daily pattern of the agent's decision-making in managing battery charge levels alongside PV generation and electricity pricing for a random day in the year. The solid blue line illustrates the battery control policy of the agent. The red and green dotted lines represent the maximum and minimum battery charge levels, with the lower limit set at 15\% and the upper threshold at 85\%. We adopt this strategy to enhance the battery's lifetime\cite{battery_university_808}. The purple dotted line illustrates fluctuations in electricity pricing, while the yellow-shaded area shows the daily generation of PV electricity and the pink-shaded area shows electricity demand in the farm. The figure shows agent behavior in charging the battery when electricity is cheaper or solar power is available and discharging it during high-price periods or when solar output is low. These results highlight the effectiveness of this research in building the optimal policy for battery management to minimize electricity import. 

Figure \ref{reward} shows the average rewards and variance for an agent during the training of the PPO algorithm over one million episodes across ten runs. The blue line represents average rewards, while the shaded area indicates the range of rewards. The x-axis shows training episodes, and the y-axis measures rewards. Initially, rewards were highly negative, indicating exploration. The agent learned from the environment as training progressed, improving its strategy. After 200k episodes, rewards stabilized with fewer fluctuations, indicating the agent's policy had reached an optimal or near-optimal level, resulting in consistent performance.


\begin{figure}[H]
\centering
\includegraphics[width=0.95\textwidth]{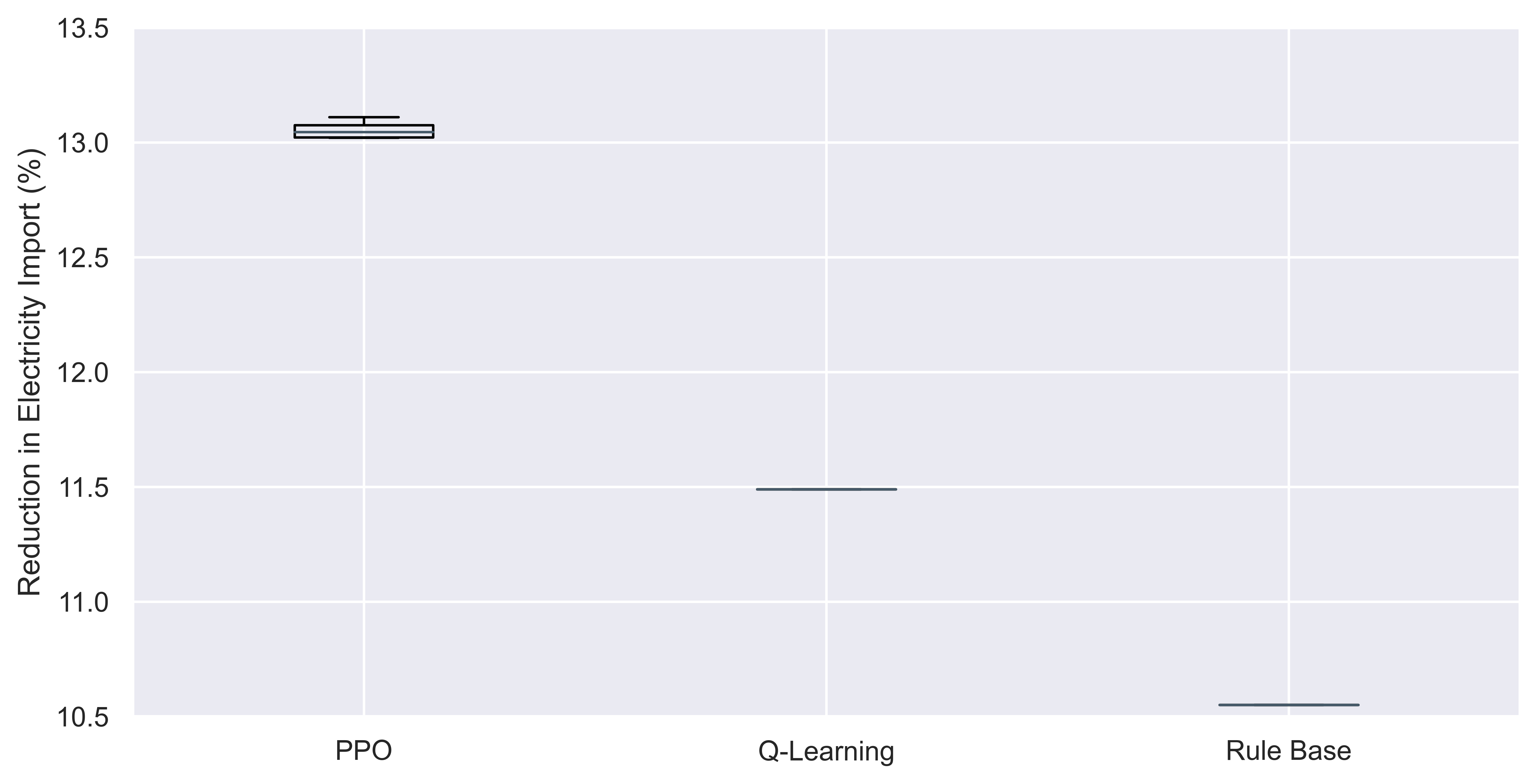}
\caption{Perforamance comaparision of algorithms.} \label{model_comparision}
\end{figure}


Figure \ref{model_comparision} shows a box plot of load reduction percentages for PPO, Q-learning, and rule-based algorithms over ten runs using 11 months of data. Rule-based and Q-learning had stable, consistent performance. PPO's stochastic policy introduced randomness, leading to varying actions in the same state across tests.

\section{Conclusion}

\begin{itemize}
\item We implemented the PPO algorithm for battery management in dairy farm settings, aiming to maximize the utilization of locally generated PV energy and reduce reliance on the electricity grid. The PPO algorithm is highlighted for its ability to make stochastic policy decisions, which allows for a more robust and adaptable decision-making process in battery management.
\item The outcome shows that the PPO algorithm for managing batteries effectively reduces the amount of electricity purchased from the grid by 13.11\% compared to scenarios with no battery, 1.62\% compared to Q-learning, and 2.56\% compared to rule-based algorithms.
\item We analyzed the algorithm's effectiveness in charging the battery when electricity prices were low or solar power was available, and discharging during high-price periods or low solar output. The results show the algorithm's efficiency in managing battery usage for dairy farms.

\end{itemize}

In future work, we plan to extend this research to include a wind generation profile to see the adaptability of the implemented algorithm. Also, we plan to test this algorithm on data from different different geographical regions and compare our work with various DRL algorithms.  

\section*{Acknowledgements} This publication has emanated from research conducted with the financial support of Science Foundation Ireland under Grant number [21/FFP-A/9040].

\bibliographystyle{unsrt}
\bibliography{biblography}

\end{document}